\pdfoutput=1
\documentclass[letterpaper, 10 pt, conference, left=1in,right=1in,top=1in,bottom=1.5in]{IEEEtran}  

\IEEEoverridecommandlockouts                              

\usepackage{graphics} 
\usepackage{epsfig} 
\usepackage{mathptmx} 
\usepackage{times} 
\usepackage{amsmath} 
\usepackage{amssymb}  

\usepackage[T1]{fontenc}
\usepackage{amsfonts}
\usepackage{booktabs}
\usepackage{siunitx}
\usepackage{caption} 
\usepackage{booktabs}
\usepackage{graphicx}
\usepackage{algorithmic}
\usepackage[ruled]{algorithm2e}
\usepackage{float}
\usepackage[percent]{overpic}

\usepackage{mwe,tikz}\usepackage[percent]{overpic}

\usepackage{subcaption}
\usepackage{array,   tabularx, makecell, booktabs}
\usepackage{geometry}
\usepackage{comment}

\title{\LARGE \bf
HabitatDyn Dataset: Dynamic Object Detection to Kinematics Estimation
}

\author{Zhengcheng Shen$^{1}$, Yi Gao$^{1}$\thanks{$^{1}$ Zhengcheng Shen , Linh K{\"a}stner, Magdalena Yordanova, and Jens Lambrecht are with the Chair Industry Grade Networks and Clouds, Faculty of Electrical Engineering, and Computer Science,				
		Berlin Institute of Technology, Berlin, Germany
		{\tt\small zhengcheng.shen@tu-berlin.de}}, Linh K{\"a}stner$^{1}$, and Jens Lambrecht$^{1}$
}

\begin{document}

\maketitle
\thispagestyle{empty}
\pagestyle{empty}


\begin{abstract}
The advancement of computer vision and machine learning has made datasets a crucial element for further research and applications. However, the creation and development of robots with advanced recognition capabilities are hindered by the lack of appropriate datasets. Existing image or video processing datasets are unable to accurately depict observations from a moving robot, and they do not contain the kinematics information necessary for robotic tasks. Synthetic data, on the other hand, are cost-effective to create and offer greater flexibility for adapting to various applications. Hence, they are widely utilized in both research and industry. In this paper, we propose the dataset HabitatDyn, which contains both synthetic RGB videos, semantic labels, and depth information, as well as kinetics information. HabitatDyn was created from the perspective of a mobile robot with a moving camera, and contains 30 scenes featuring six different types of moving objects with varying velocities. To demonstrate the usability of our dataset, two existing algorithms are used for evaluation and an approach to estimate the distance between the object and camera is implemented based on these segmentation methods and evaluated through the dataset. With the availability of this dataset, we aspire to foster further advancements in the field of mobile robotics, leading to more capable and intelligent robots that can navigate and interact with their environments more effectively. The code is publicly available at https://github.com/ignc-research/HabitatDyn.
\end{abstract}


\section{Introduction}
\noindent
Dynamic object detection and tracking is an essential ability to realize a safe autonomous navigation for mobile robots. By detecting the object with distance and velocity estimations, the mobile robot can avoid potential collisions in an earlier stage, especially when the environment is highly dynamic \cite{DBLP:journals/ijrat/RaiT15} \cite{DBLP:journals/corr/abs-2109-11288}. Dynamic detection algorithms can also improve the robustness of a SLAM algorithm \cite{DBLP:journals/corr/abs-2109-13160}. The further interaction and tracking ability also rely on the accuracy of the dynamic object detection\cite{doi:10.1142/7017}. Although research in dynamic object detection is making great progress thanks to the recent advances in deep learning algorithms, the limitation of the dataset and incomparable evaluation processes make it hard to develop, train, and evaluate algorithms \cite{7780454}. When it comes to detecting moving objects in indoor environments, one major challenge is obtaining real-world data. In addition to requiring a significant amount of data, the datasets need to be tailored to the specific task, such as differentiating between the behaviors of pedestrians or agents in an indoor environment. As a result, dynamic object detection algorithms often lack the ability to extend to mobile robots, and further training and adaptations are necessary for deployment on real robots.
\begin{figure}[!ht]
\centering
\includegraphics[width=0.45\textwidth]{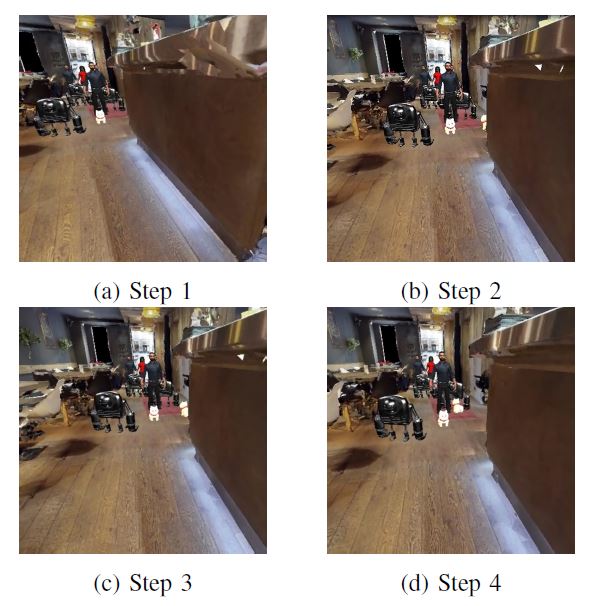}
\caption{The dataset is divided into two categories: one with only dynamic objects and the other with objects that can be either static or dynamic. These four snapshots from the second category show a moving dog appearing in the second step, while the other objects remain static. The camera movement also changes the perspective of the static objects in the background.}
\label{fig:introduction}
\end{figure}
\noindent To address these challenges, this paper presents: 
\begin{itemize}
    \item A synthetic dataset that includes RGB videos on a moving camera, semantic labels, depth information, and object locations. 

    \item Evaluation of two dynamic objects detection algorithms for videos taking on the moving camera on the new dataset

    \item A workflow for distance estimation of the multiple dynamic objects with a moving camera based on a segmentation algorithm and a clustering algorithm is implemented and evaluated. 

\end{itemize}
The dataset was created from the perspective of a mobile robot with a moving camera and includes 30 scenes featuring six different types of moving objects with varying velocities, and aims to provide the necessary data to facilitate future research in dynamic object detection and enhance robot functionality. Figure \ref{fig:introduction} displays four sequential example views from one of the videos, illustrating how the objects are organized based on their static or dynamic status.

\noindent The paper is structured as follows. Sec. II begins with related
works followed by our problem statement. Subsequently, the
methodology is presented in Sec III. The details about the
implementation are described in Sec. IV. Sec. V presents the
results and discussion. Finally, Sec. VI will give a conclusion
and outlook.

\section{Related Works}
\noindent With the advances in object detection and instance segmentation methods, research in navigation within dynamic environments became increasingly popular in recent years. A majority of the proposed approaches for SLAM in dynamic environments or collision avoidance algorithms partially rely on ta deep learning model. In \cite{kiran2019realtime}, \cite{DBLP:journals/corr/abs-2010-00052}, the instance segmentation algorithms based SLAM algorithms for dynamic environments are introduced. With the commonly used Mask-RCNN approach \cite{DBLP:journals/corr/HeGDG17}, the algorithm can filter out the ORB features\cite{6126544} on the dynamic objects, which leads to a stable mapping and localization. The research is based on the dataset for autonomous driving and evaluated for outdoor usage. Another research form Zhang et al. 2021 \cite{DBLP:journals/corr/abs-2005-11052} argue that the dynamic information hold potential benefits for SLAM, and it is crucial for robotics to understand dynamic environments for tasks such as planing, control and obstacle avoidance. While most of the research works focus on outdoor environment, it is also essential for robots to understand indoor dynamic environments to facilitate a range of application such as service robots in hospitals, offices, or warehouses.

\noindent For an in-door dynamic environment, Vincent et al.2020 \cite{DBLP:journals/corr/abs-2008-00072} deployed an instance-segmentation-based approach to extend the Kalman filter and improve the accuracy and stability of a visual SLAM. DynaSLAM\cite{DBLP:journals/corr/abs-1806-05620} is also a visual SLAM approach proposed by Bescos te al. to facilitate robust navigation in Dynamic Scenes. Since the community oftentimes focuses on the potential solution for dynamic environments and dynamic object detections, a dataset based on dynamical environment is essential for further research, training, evaluation, and deployment of a deep-learning-based algorithm. The current dynamic object datasets like Davis\cite{7780454} or YouTube-VOS 2018 \cite{xu2018youtubevos} are useful for training of the deep learning methods but not ideal for more kinematics estimation and evaluation, like pose and orientation estimation, velocity estimation, future movement estimation, or distance estimation. On this account, this work introduces a method, which is a combination of dynamic object segmentation algorithms, to estimate the kinematics of the dynamic object in an indoor environment. Furthermore, a dataset with full diversity and specialized for mobile robot application development is introduced.
\begin{figure}[!htp]
	\centering
	\includegraphics[width=0.45\textwidth]{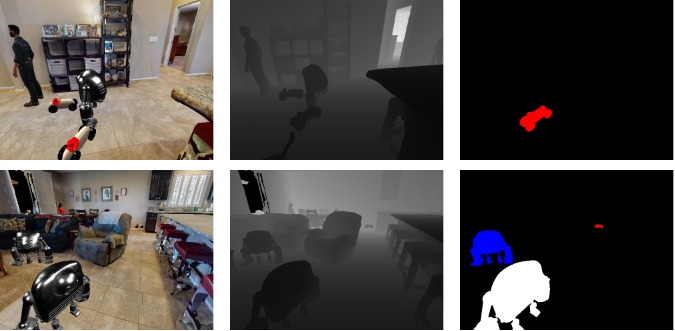}
	\caption{The dataset contains several visual information including RBG data, depth image and moving object semantic annotations for different objects. Although the object labels are in different colors, we also provide the object ID labeling from 1 to 6. Furthermore, there are location information, orientation information, camera coefficient and maximum movement speed information for kinematics research with advanced deep learning methods.}
	\label{fig:data_samples}
\end{figure}

\section{Dataset Description}
\noindent In this section, the dataset will be introduced and explained. The dataset is carefully constructed in the way that take care of the demands from combination of machine learning based segmentation algorithm and mobile robot application purpose. The entire data folder struct can be similar with Davis\cite{7780454}, which make us easier to

\subsection{Data Amount and Quality}
\noindent The quality and quantity of data are crucial factors, as they directly impact the performance, accuracy, and generalizability of the models being developed. The dataset comprises 720 short clips, captured by an embodied agent within a dynamic environment created using Habitat-sim. Each clip lasts 10 seconds and is recorded at a frequency of 24 frames per second (fps), resulting in 241 frames per clip. In total, there are 173,520 annotated frames, providing a substantial amount of data for training and evaluating the proof of concept for the algorithm. The data quality is tailored to the current onboard processing capacity of a mobile robot, utilizing a 480p resolution. This choice not only meets the robot's processing capabilities, but also enhances the efficiency of the training and evaluation processes.

\subsection{Data Diversity}
\noindent The dataset aims to provide sufficient diversity to effectively evaluate various approaches across multiple dimensions. It consists of recordings from 30 distinct scenes based on the photorealistic MP3D dataset, captured in real apartments. This variety ensures a diverse range of backgrounds, including different types of furniture and wall or floor textures. It also narrows the gap between synthetic data and reality compared to other synthetic datasets. SLAM research based on Habitat-sim can be transitioned to real-world environments with minimal effort \cite{DBLP:journals/corr/abs-2004-05155}.

\noindent The 720 clips are divided into two categories: pure dynamic clips and mixed dynamic clips. In the pure dynamic clips category, all embedded objects are dynamic, while only the background from MP3D remains static. To prevent overfitting for object detection rather than dynamic object detection, the mixed dynamic clips category contains both static and dynamic embedded objects. As illustrated in figure \ref{fig:introduction}, a static dog appears in front of a human while another dynamic dog moves around the corner.

\noindent The dataset contains six moving objects, with half of the videos featuring only one type of dynamic object and the other half featuring a mix of three different dynamic objects. This enables a comparison of segmentation results across various objects and scenarios. Additionally, half of the clips contain three objects, while the others feature six moving objects in total.

\noindent All videos are divided into three speed categories, with each clip featuring objects moving with the same maximum speed. In the turning around action, the objects will decrease the speed. This allows for an assessment of the algorithm's segmentation ability at different speed levels without losing generality.

\subsection{Object Presence}
\noindent The object presence is different in mobile robot navigation compared to other applications. Most of the existing datasets contain the objects in every frame of the video, which may lead the model to believe that every frame contains the dynamic object and be overconfident for the estimation. Since the dynamic objects are randomly distributed and are performing a repetitive moving strategy according to our path setting, some of the frames may not include any moving object and some of them may have all the objects. The randomness of the presence is a more realistic way to handle the object presence.

\subsection{Kinematics}
\noindent To bring the concept of the kinematics into the dataset, we integrate the depth information, the global coordinates of the camera and the objects, and the orientation of the objects into the dataset. With the extension of those extra data and the high sample frequency, the dataset can be used for distance estimation evaluation, speed estimation, future movement prediction, object pose/orientation estimation. The dataset extends the normal computer vision dataset with essential mobile robot research requirements. Further parameters for the kinematic calculation are listed in table:
\begin{table}[!h]
\begin{center}
\begin{tabular}{ | m{3cm}| m{3cm} | } 
  \hline
  Camera HFoV & 90 degree \\
  \hline
  Sensor Height & 1.25 meter\\
  \hline
  Frequency & 24Hz \\
  \hline
  Max Speed & 1 m/s, 2 m/s, 3 m/s \\
  \hline
  Max Depth & 10 meter \\
  \hline
  Resolution & 640x480 \\
  \hline
  Object ID & 1 - 6 \\
  \hline
\end{tabular}
\end{center}
\caption{Hyperparameter details}
\label{table:4_hyperp}
\end{table}
\begin{figure*}[!ht]
	\centering
	\includegraphics[width=0.8\textwidth]{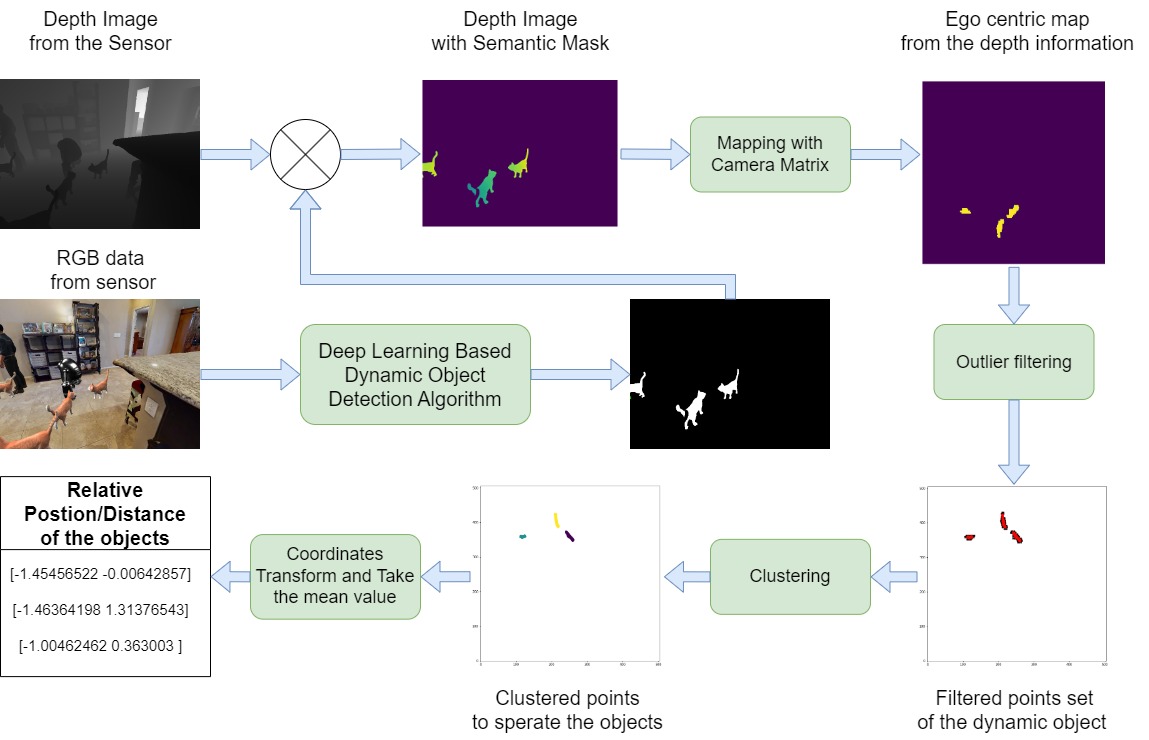}
	\caption{Overview of the experiment architecture. We use the Gibson scenes with habitat-sim for the training and evaluation. Besides the original ANS algorithm, the dynamic object module, the annotation of the dynamic object, and an adapt mapper unit with different approaches are required for the extension. With the RGB observation and the optimized mapper module as input, the ANS module is responsible for the SLAM and exploration task.}
	\label{system}
\end{figure*}
\section{Evaluated Algorithms}
\noindent Moving object detection in videos, also be referenced as salient objects detection, aims to label all interesting object(s) in consecutive video frames as foreground, regardless of which object category, e.g. dog, cat, it belongs to. So the understanding of the temporal relation of in-frame objects is critical. Conducting instance segmentation in video frames makes it even more challenging, as the model needs to perform dense per-pixel interpretation and requires a large receptive field concerning each pixel. On top of that, multi-scale features are also crucial to do per-pixel prediction, for instance segmentation. Common backbone structures that can leverage the multi-scale features includes Feature Pyramid Networks\cite{lin2017feature}, ResNet\cite{DBLP:journals/corr/HeZRS15}, U-net\cite{DBLP:journals/corr/RonnebergerFB15}  and Swin-transformer\cite{DBLP:journals/corr/abs-2103-14030}. 

\subsection{Dynamic Object Detection}
\noindent The term unsupervised can be ambiguous in today’s deep learning field, unsupervised models mostly indicate that the models do not need any “ground truth” for training, making it highly appealing as it doesn’t need the effort for labeling the data, like the CIS model \cite{DBLP:journals/corr/abs-1901-03360}. As for the video dataset DAVIS 2016 unsupervised\cite{7780454}, the term “unsupervised“ is referring to the fact that during inference, the model does not need the user’s interaction to suggest the interesting or salient object for tracking. 

\noindent Two different algorithms, namely, 3DC-seg\cite{Mahadevan20BMVC} and CIS, are used to evaluate the datasets generated by this work. While 3DC-seg is the leading model in the DAVIS 2016 unsupervised dataset benchmark, the CIS model from 2019 ranks at around 20th place of the benchmark, it still has illuminating points like it does not acquire any ground truth annotation and thus can be seen as a truly unsupervised model.

\subsection{Kinematics Estimation}
\label{Kinematics}
\noindent With the output of the dynamic object detection algorithm, we can make the kinematics estimation within the HabitatDyn dataset. The figure \ref{system} demonstrates the algorithm workflow in a better visualized way. The algorithm takes depth information $obs_{depth}$ and RGB data $obs_{rgb}$for the input, and the output is the coordinates' estimation $(x,y)$ in the coordinate frame of the robot for the moving object. First, the mask for the moving object $m_{moving}$ is generated by the deep-learning-based dynamic detection algorithm. Subsequently, the masked depth observation will then be calculated by $obs_{depth_masked} = obs_{depth}*m_{moving}$. To further segment and process $obs_{depth_masked}$, we map it first into the point clouds with inverse the intrinsic camera matrix and then map the points again into a top-down view $map_{ego}$, where we have to dine the map resolution $r$. Since the limitation of segmentation, the map will produce outliers from the flaw of the segmentation on the edge, even with high accuracy ground truth labeled data. Therefore, a filter for outliers based on the Identifying-Density-Based-Local-Outliers approach \cite{inproceedings} is implemented to reduce the interference and prepare the data for clustering. Before the outlier filtering, we transfer the data presentation of the $map_{ego}$ again into two-dimensional points arrays $P$ by $P = {(x, y) \mid \text{map}_{\text{ego}}(x, y) > 0}$
Afterward, the points will be classified into objects with the DBSCAN approach \cite{10.5555/3001460.3001507} and coordinates will be calculated accordingly with a rescaling factor $r$ and a translation from map coordinates to robot coordinates.
\begin{algorithm}
\caption{Coordinates Estimation within HabitatDyn Dataset}
\label{alg:kinematics_estimation}
\begin{algorithmic}[1]
    \REQUIRE Depth information $obs_{depth}$, RGB data $obs_{rgb}$
    \ENSURE Coordinates' estimation $(x, y)$ of moving object in the robot's coordinate system
    \STATE Generate the mask for the moving object $m_{moving}$ using a deep learning-based dynamic detection algorithm
    \STATE Compute masked depth observation: $obs_{depth\_masked} = obs_{depth} \cdot m_{moving}$
    \STATE Map $obs_{depth\_masked}$ to cloud points using inverse intrinsic camera matrix
    \STATE Map cloud points to a top-down view $map_{ego}$ with a defined map resolution $r$
    \STATE Convert $map_{ego}$ data to a two-dimensional points array $P$: $P = \{(x, y) \mid \text{map}_{\text{ego}}(x, y) > 0\}$
    \STATE Apply a filter for outlier removal based on Identifying Density-Based Local Outliers\cite{inproceedings}
    \STATE Classify points into objects using the DBSCAN algorithm\cite{10.5555/3001460.3001507}
    \STATE Calculate coordinates $(x, y)$ in the robot's coordinate system using rescaling factor $r$ and appropriate translation
\end{algorithmic}
\end{algorithm}

\begin{figure*}[!htp]
	\centering
	\includegraphics[width=0.95\textwidth]{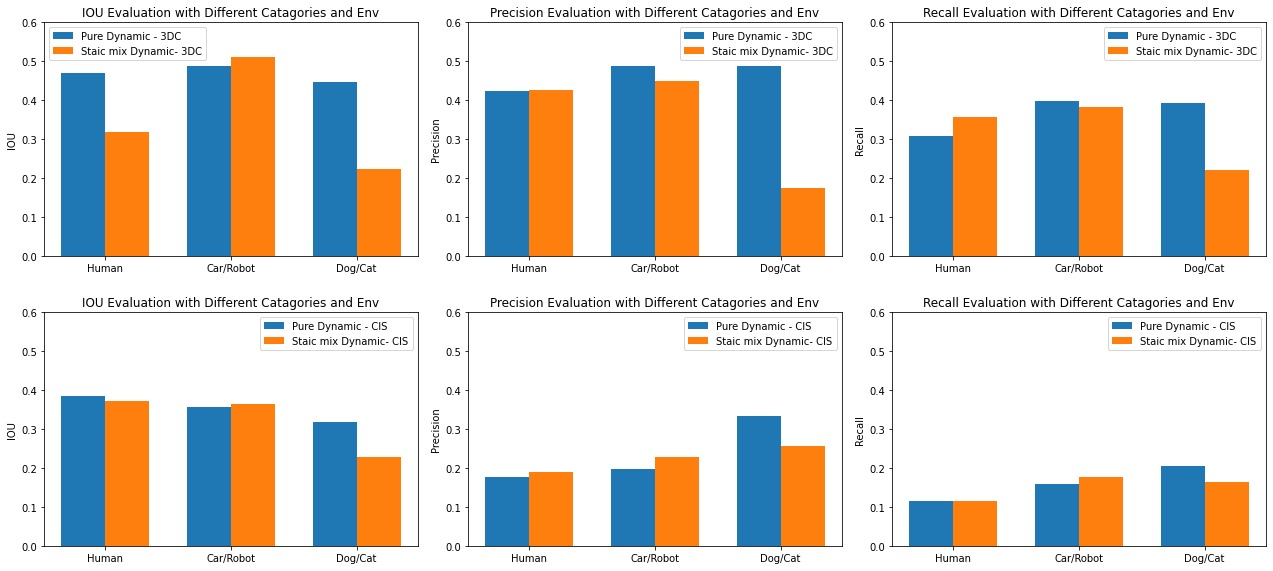}
	\caption{Evaluation of the algorithm within different categories in two different environment. The image shows 6 bar charts to compare the performance of two dynamic object detection algorithms within different categories. The performance changing is quite different in different semantic groups.}
	\label{fig:eval_categories}
\end{figure*}
\begin{figure}[!htp]
	\centering
	\includegraphics[width=0.4\textwidth]{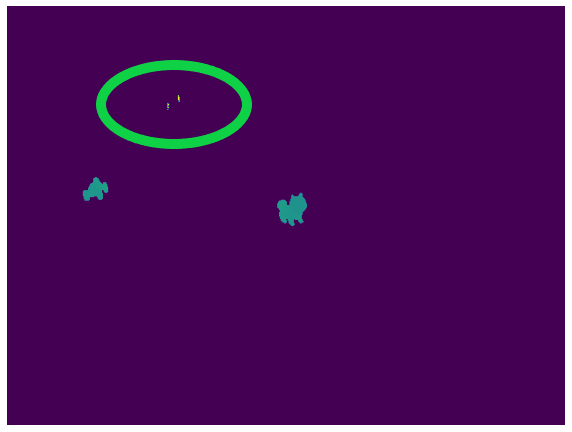}
	\caption{There are three objects in the label. But the object in the green cycle is too far and too small to detect with low resolution sensors}
	\label{fig:far_object}
\end{figure}
\section{Implementation and Evaluation}
\noindent In this section, we will first introduce the implementation details about our dynamic objects coordinates estimation algorithm. The evaluation will be separated into two parts, the first part focuses on the evaluation of the dynamic detection algorithm. The second part evaluates the coordinates' estimation algorithm with the matrix, distance between the objects and the robots.

\subsection{Implementation}
\noindent In section \ref{Kinematics}, we have introduced the algorithm in detail. The aglorith we used for generating the mask is based on 3DC-seg. The map size we used in the implemention is 1000 pixels width and 1000 pixels high. The resolution of the map against the real world is 0.01 meter between each two pixels. This means the maximum detection distance of the algorithm is 10 meters. Beyond the distance, the object will not show up on the egocentric map $map_{ego}$. The coefficient number of neighbors we used in LOF and DBSCAN is set up as $n_{neigbors} = total points number/30 + 1$, the contamination of the LOC is 0.3, which have a balance to eliminate the outliers but keep small objects being discovered.

\subsection{Evaluation of Dynamic Object Detection Algorithm}
\begin{table*}[!ht]
    \centering
    \caption{Algorithm Evaluation with Pure Dynamic Embedded Objects and Mixed Embedded Objects}
    \label{tab:staitc and dynamic}
    \begin{tabular}{llllll}
        \toprule
        Algorithm & Single class, excl. static  & Multi-class,excl. static & Single class, incl. static & Multi-class, incl.static & Avg \\
        \midrule
        & \multicolumn{5}{c}{IOU} \\
        \cmidrule{2-6}
        CIS & 0.3666 & 0.369 & 0.363 & 0.323 & 0.355\\
        3DC & 0.515 & 0.508 & 0.294 & 0.331 &  0.413\\
        \midrule
        & \multicolumn{5}{c}{Precision} \\
        \cmidrule{2-6}
        CIS & 0.244 & 0.237 & 0.211 & 0.196 & 0.223\\
        3DC & 0.522 & 0.504 & 0.237 & 0.263 & 0.387\\
        \midrule
        & \multicolumn{5}{c}{Recall} \\
        \cmidrule{2-6}
        CIS & 0.171 & 0.1546 & 0.158 & 0.131 & 0.153\\
        3DC & 0.421 & 0.382 & 0.257 & 0.320 & 0.346\\
        \bottomrule
    \end{tabular}
\end{table*}
\subsection{Evaluation of Dynamic Object Detection Algorithm For Different Obejct Movement Speed}
\begin{table*}[!ht]
    \centering
    \caption{Algorithm Evaluation with Pure Dynamic Embedded Objects and Mixed Embedded Objects}
    \label{tab:movement Speed}
    \begin{tabular}{lllllll}
        \toprule
        Algorithm & speed1\_excl.static & speed2\_excl.static & speed3\_excl.static & speed1\_incl\_static & speed2\_incl\_static & speed3\_incl\_static \\ \hline
        \midrule
        & \multicolumn{6}{c}{IOU} \\
        \cmidrule{2-7}
        CIS & 0.362 & 0.350 &  0.345 & 0.356 & 0.301 & 0.300\\
        3DC & 0.505 & 0.493 &  0.506 & 0.292 & 0.316 & 0.266\\
        \midrule
        & \multicolumn{6}{c}{Precision} \\
        \cmidrule{2-7}
        CIS & 0.276 & 0.196 & 0.219 & 0.148 & 0.223 & 0.192\\
        3DC  & 0.526 & 0.493 &  0.508 &  0.262 & 0.237 & 0.247\\
        \midrule
        & \multicolumn{6}{c}{Recall} \\
        \cmidrule{2-7}
        CIS & 0.192 & 0.126 & 0.141 & 0.106 & 0.117 & 0.130\\
        3DC & 0.417 & 0.394 & 0.409 & 0.281 & 0.246 & 0.195\\
        \bottomrule
    \end{tabular}
\end{table*}
We evaluate the two algorithms, 3DC-seg and CIS, with our dataset in different perspectives. The matrix we used in the evaluation is IOU, Precision and Recall. The IOU is calculated according to:
\begin{equation}
    \text{IOU} = \frac{\text{Area of Intersection}}{\text{Area of Union}} = \frac{|A \cap B|}{|A \cup B|}
\end{equation}
where A and B represent the estimated mask and the ground truth mask.
The Precision and Recall values are calculates as following:
\noindent 
\begin{equation}
    \text{Precision} = \frac{\text{TP}}{\text{TP + FP}}
\end{equation}
\begin{equation}
    \text{Recall} = \frac{\text{TP}}{\text{TP + FN}}
\end{equation}
where TP is the true positive, FP is the false negative and FN is the false positive. The precision measurement is a common measurement for segmentation. High precision means that, among the instances predicted as positive, a high proportion is actually positive. Recall is used to measure if the algorithm has the ability of a classifier to identify all the positive instances.

\noindent The table \ref{tab:staitc and dynamic} show the evaluation results of these two methods for the environment with dynamic embedded objects and the environment with both dynamic and static embedded objects. The performance of both algorithms drops when the static embedded objects are introduced. The decline of the performance is less on 3DC, but the algorithm also performs not well in a dynamic-embedded-object-only scenario. The IOU and Precision drops significantly because it takes a a high number of static embedded objects as dynamic object. The figure \ref{fig:eval_categories} shows the performance of the two methods in different categories. The performance decline of 3DC between those two environments on cat\\dog category is significant and within every matrix. It may be caused by the outstanding color of skin of the 3D model, since the method is used for foreground segmentation. The human category also has the lowest Recall measurement in both of the methods. One possible reason could be that the Human class has a larger size than other models and has different parts such that it is harder to generate a mask to cover the entire body. The table \ref{tab:movement Speed} shows the influence of the object movement speed on the algorithm. The speed increase of the moving object has only minor influence on the performance of 3DC but has substantial influence on the CIS algoritm.

\subsection{Evaluation about Distance Estimation}
\begin{figure*}[!htp]
	\centering
	\includegraphics[width=0.95\textwidth]{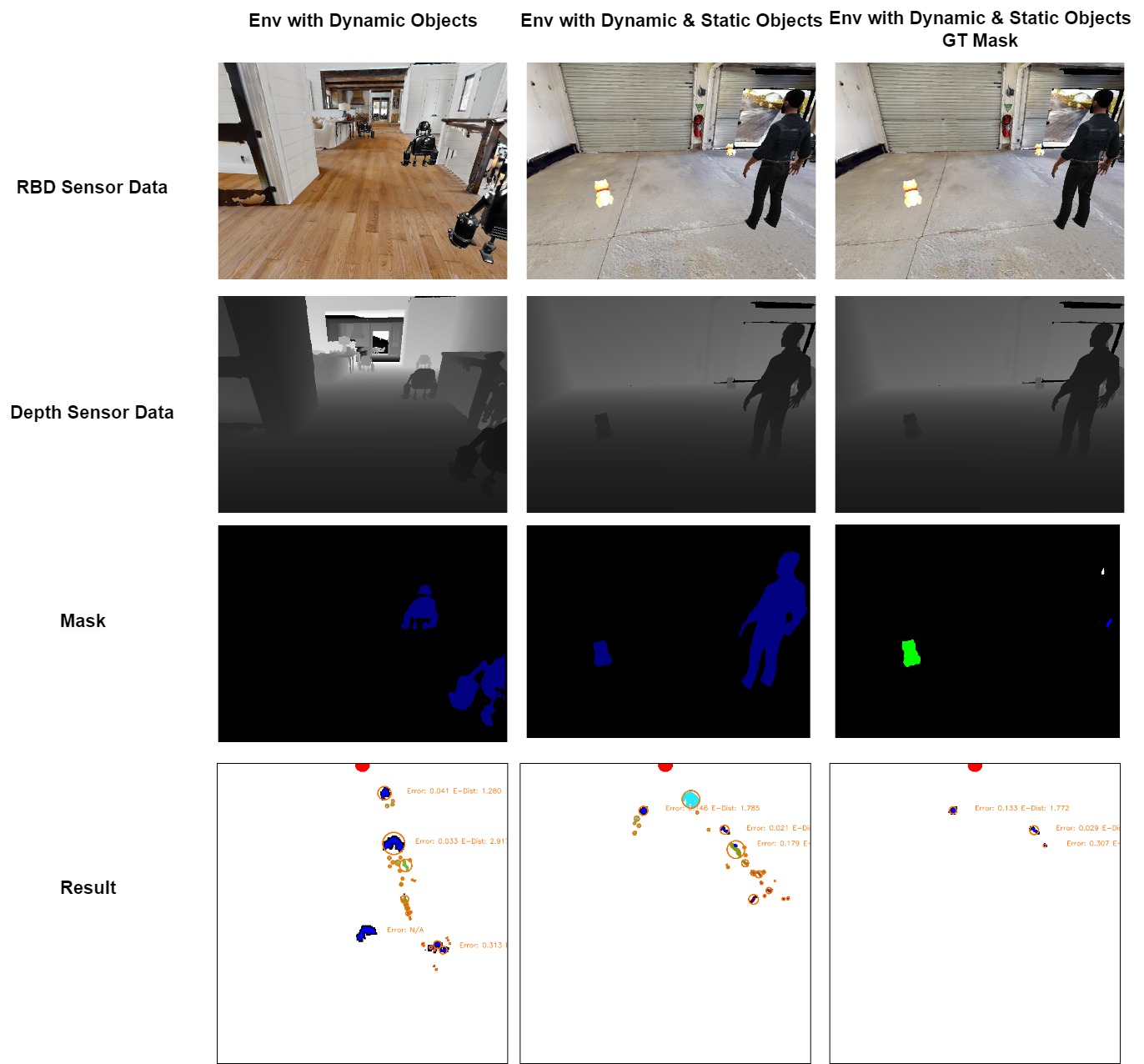}
	\caption{The images show the sensor data and the three examples from the evaluation scenes. The three columns are corresponding to samples from dynamic-objects-only environment, environment with dynamic and static embedded objects but with predicted mask and with ground truth mask. The red dot represents the robots' position. The blue dots represent the points from ground truth. the colorful dots represent the location of the predicted moving objects. The center of the cycle is the estimated position for the corresponding objects, and radius is the max distance from the points in the cluster to the center of the cycle.}
	\label{fig:dis_result}
\end{figure*}
\begin{table*}[!ht]
    \centering
    \caption{The Detection Rate, Threshold Accuracy and MAE for Different Distance}
    \label{tab:distance}
    \begin{tabular}{|l|l|l|l|l|l|l|l|l|}
    \hline
        Distance(m) & Det Rate & MAE & SquaRel & RMSE & RMSLE & $\delta$ & $\delta^2$ & $\delta^3$\\ \hline
        (0, 3] & 91.22\% &0.201&0.196  & 0.407 & 0.173 & 92.90\% & 97.87\% & 99.26\%\\ \hline
        (3, 6] & 62.03\% & 0.078 &0.010 &0.092 &0.046 & 100\% & 100\% & 100\% \\ \hline
        (6, 10] & 0.09\% & 0.051 &0.005 &0.066 &0.033 & 100\% & 100\% & 100\%\\ \hline
    \end{tabular}
\end{table*}
\noindent After calculation of the coordinates in the mobile robot's coordinate frames, we leverage the ability of the dataset and evaluate the distance estimation function for dynamic objects. The distance between the robot and the dynamic object can be useful for navigation \cite{DBLP:journals/corr/abs-2109-11288}. The estimated distance will be calculated by $d = \sqrt{x^2 + y^2}$, where x and y are the estimated coordinates to represent the moving objects.
To evaluate the performance of the function, we face several problems:
\begin{itemize}
    \item Not all the moving object can be detected by the algorithm.
    \item The algorithm may detect more moving objects than it has in the current see sight, which is called illusion or overconfidence.
    \item The detected moving object is not related to ant label ID we have.
\end{itemize}
To evaluate the ability to detect and calculate the distance, we follow the several conditions. We considered objects, which have 10 meter or less distance to the robot, since the simulator also has a limited maximum depth of 10 meters. For mobile robots, an object 10 meters away  has also minor influence to its navigation strategy. The estimation will first use the label to find the objects in the field of view and map the depth image into a top-down view. Subsequently, an intersection will be performed between the cluster from the labeled data and the cluster from the estimated data. A match will be established if two have the maximum intersections among all the pairs in regard to this frame.
For the data pairs, the following measurements will be used:
\begin{equation}
    \mathrm{MAE} = \frac{1}{n} \sum_{i=1}^{n} |d_i - \hat{d}_i|
\end{equation}
\begin{equation}
    \text{SquaRel}_i = \frac{1}{n}\sum_{i=1}^{n}\left(\frac{d_{i} - \hat{d}_{i}}{d_{i}}\right)^2
\end{equation}
\begin{equation}
    \text{RMSE} = \sqrt{\frac{1}{n}\sum_{i=1}^{n}(d_{i} - \hat{d}_{i})^2}
\end{equation}
\begin{equation}
    \text{RMSLE} = \sqrt{\frac{1}{n}\sum_{i=1}^{n}\left(\log(1+d_{i}) - \log(1+\hat{d}_{i})\right)^2}
\end{equation}
where n is the number of samples, $d_i$ represents the true value of the distance, and $\hat{d}_i$ represents the predicted distance.
We also use the threshold accuracy matric:
\begin{equation}
   \text{Threshold} = \%\ of\ d_i s.t.max(\frac{\hat{d}_i}{d_i},\frac{d_i}){\hat{d}_i} = \delta < \text{threshold})
\end{equation}
where the threshold usually takes the value: $\delta < 1.25, \delta<1.25^2,\delta<1.25^3 $.

\noindent Once the match is found, the estimation will take on this exact pair of data. Table \ref{table:result1_dis} shows that 74.64\% of the objects from the label are matched with a detected cluster and the estimation of the threshold accuracy. Although the environment including static embedded objects reduce the performance of the dynamic object detection algorithm, the overall coordinates and distance estimation takes only a slight decrease of the detection rate.In figure \ref{fig:dis_result}, we demonstrate the samples of the results, the closed objects are mostly being detected while we lost the mask of the objects behind, since the 3DC is used to detect foreground objects. In the second column, we can find the static man is also being masked. It causes the wrong detection of dynamic objects, but whether it has a positive or negative influence on safety navigation is to be discussed and to be researched. In table \ref{tab:distance}, we evaluate the approach according to different distance interval. The detection rate drops rapidly when the distance goes larger. But on the other hand, the error cause by the distant objects is much smaller. Since the distant object occupied only a small area of the pixel. An example is provided in figure \ref{fig:far_object}. It has the possibility to be removed by the algorithm as an outlier. The large error in the nearest area is due to the limitation resolution of the map. A possible solution is to create a gradient mesh to make the resolution of the map higher in the near robot area. The center point calculation can also be optimized for future performance improvement.

\begin{table}[!ht]
    \centering
    \caption{The Detection Rate and Threshold Accuracy}
    \label{table:result1_dis}
    \begin{tabular}{|l|l|l|l|l|}
    \hline
        Environment & Det Rate & $\delta$ & $\delta^2$ & $\delta^3$\\ \hline
        Pure Dynamic & 74.64\% & 92.90\% & 97.87\% & 99.26\%\\ \hline
        Static mix Dynamic & 71.87\%  & 93.29\% & 98.31\% & 99.45\% \\ \hline
    \end{tabular}
\end{table}

In this paper, we presented the dataset HabitatDyn, which contains video records, depth information, and semantic labeling for dynamic environments. We were able to evaluate two dynamic object detection algorithms, CIS and 3DC, in multiple aspects and demonstrated that the two evaluated methods can be strongly influenced by static foreground objects. CIS will also be influenced by the speed of the
object, while 3DC is not sensitive to it. Based on the dynamic object detection algorithm, we also proposed a dynamic object distance estimation algorithm and evaluate it with HabitatDyn. The approach show a stable performance within a short distance, despite the volatile performance from the dynamic object detection algorithms. The approach shows that it is important to have a dataset as a bridge and
enable the research to test and evaluate different approaches. Ongoing works aim to refine the dataset for more mobile robotic related task and to refine the dynamic object detection algorithm by adding velocity estimation, pose estimation, movement prediction, the stability and detection rate of the
algorithm. Furthermore, we aspire to extend the robustness of the dynamic object detection with idle movable objects.


\addtolength{\textheight}{-1cm} 




\typeout{}
\bibliographystyle{IEEEtran}
\bibliography{ref}

\end{document}